\documentclass[10pt,twocolumn,letterpaper]{article}

\usepackage[pagenumbers]{wacv} 

\definecolor{wacvblue}{rgb}{0.21,0.49,0.74}
\usepackage[pagebackref,breaklinks,colorlinks,allcolors=wacvblue]{hyperref}
\usepackage{algorithm}
\usepackage{algorithmic}
\usepackage{xcolor}
\usepackage{graphicx}
\usepackage{booktabs}

\usepackage{multirow}
\usepackage{multicol}
\usepackage{makecell}
\usepackage{float}
\usepackage{amsmath}
\usepackage{url}
\usepackage{bm}
\usepackage{pifont}
\def\wacvPaperID{1080} 
\def\confName{WACV}
\def\confYear{2026}

\title{VLMs Guided Interpretable Decision Making for Autonomous Driving}

\author{Xin Hu$^{1}$\thanks{This work is partially supported by NIH 2U19 AG055373-06A1 as well as the Harold L. and Heather E. Jurist Center of Excellence for Artificial Intelligence at Tulane University.}, Taotao Jing$^{2}$, Renran Tian$^3$, Zhengming Ding$^{1}$\\
$^1$Department of Computer Science, Tulane University  $^2$Qualcomm\\
$^3$Department of Industrial and Systems Engineering, North Carolina State University\\
}

\begin{document}
\maketitle
\begin{abstract}
Recent advancements in autonomous driving (AD) have explored the use of vision-language models (VLMs) within visual question answering (VQA) frameworks for direct driving decision-making. However, these approaches often depend on handcrafted prompts and suffer from inconsistent performance, limiting their robustness and generalization in real-world scenarios. In this work, we evaluate state-of-the-art open-source VLMs on high-level decision-making tasks using ego-view visual inputs and identify critical limitations in their ability to deliver reliable, context-aware decisions. Motivated by these observations, we propose a new approach that shifts the role of VLMs from direct decision generators to semantic enhancers. Specifically, we leverage their strong general scene understanding to enrich existing vision-based benchmarks with structured, linguistically rich scene descriptions. Building on this enriched representation, we introduce a multi-modal interactive architecture that fuses visual and linguistic features for more accurate decision-making and interpretable textual explanations. Furthermore, we design a post-hoc refinement module that utilizes VLMs to enhance prediction reliability. Extensive experiments on two autonomous driving benchmarks demonstrate that our approach achieves state-of-the-art performance, offering a promising direction for integrating VLMs into reliable and interpretable AD systems.
\end{abstract}
    
\section{Introduction}
\label{sec:intro}

\begin{figure}[t]
    \centering
    \includegraphics[width=\linewidth]{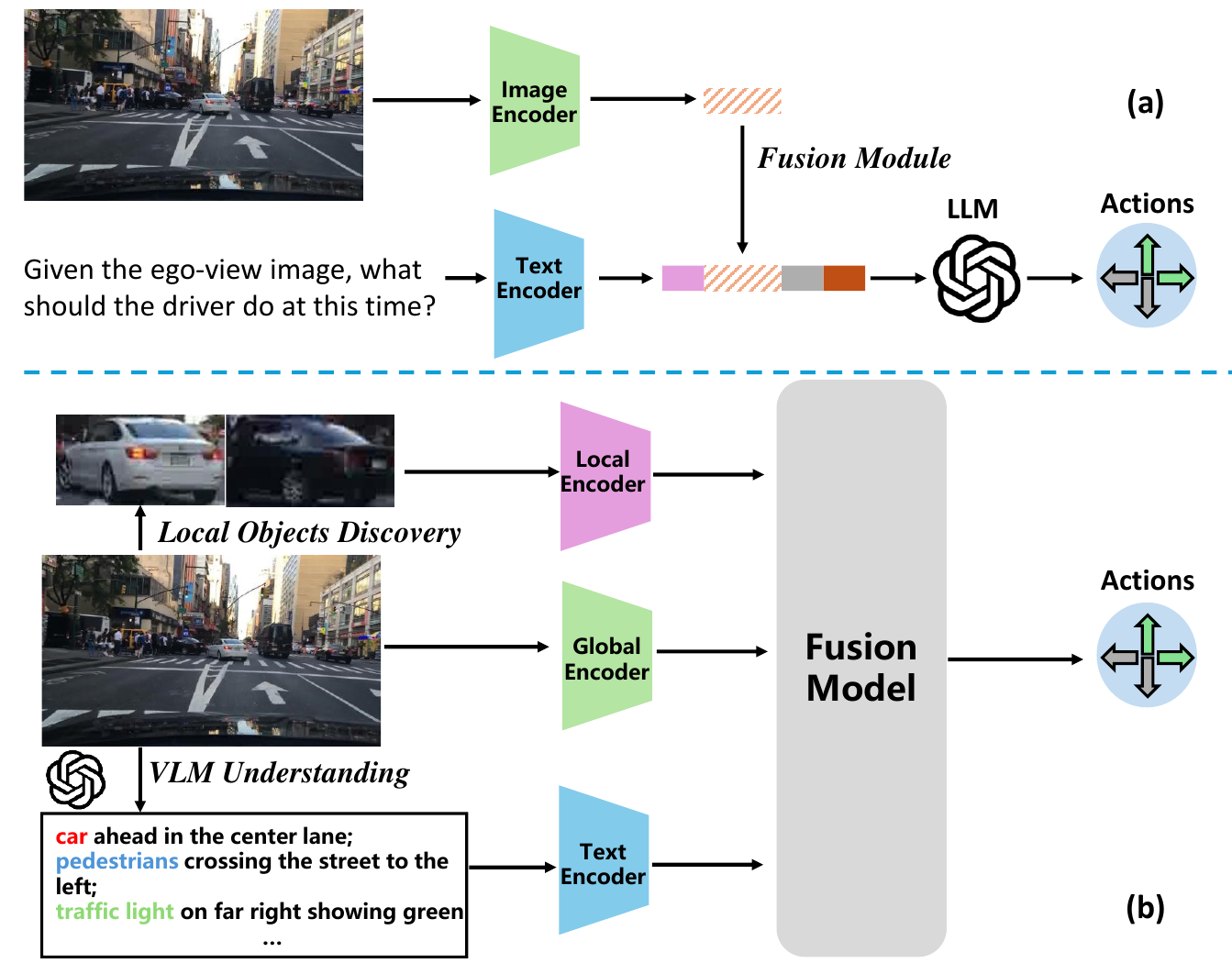}
    \vspace{-6mm}\caption{(a) illustrates the conventional VLM method for decision making, which ignores the intra-relationship between local objects, and the inter-modality alignment between local objects and text description. We propose a framework exploring both intra- and inter-modality, as depicted in (b).}\vspace{-6mm}
    \label{method-comparison}
\end{figure}

Over the past decades, autonomous driving has made significant strides, particularly with the integration of deep neural networks for end-to-end decision-making in complex driving scenarios \cite{chib2023recent}. Despite these advances, a major challenge remains: the ``black-box'' nature of these models, which makes it difficult to interpret decisions or understand the rationale behind predictions. This lack of transparency poses a critical barrier to broader AI adoption, especially in safety-sensitive domains such as healthcare and autonomous driving, where trust in decision behavior is paramount. The recent emergence of large vision-language models (VLMs), such as LLaVA \cite{liu2024visual, liu2024improved} and GPT-4V \cite{ouyang2022training}, presents a promising direction for addressing this issue by enabling natural language explanations of model behavior. These models exhibit remarkable capabilities in visual understanding and human-like language generation and have catalyzed progress across a range of research areas, including computer vision, natural language processing, and multimodal learning \cite{chang2024survey, narayanan2023multi, wang2018glue, vaswani2017attention, wang2019superglue}.

\begin{figure*}[htbp]
    \centering
    \includegraphics[width=\linewidth]{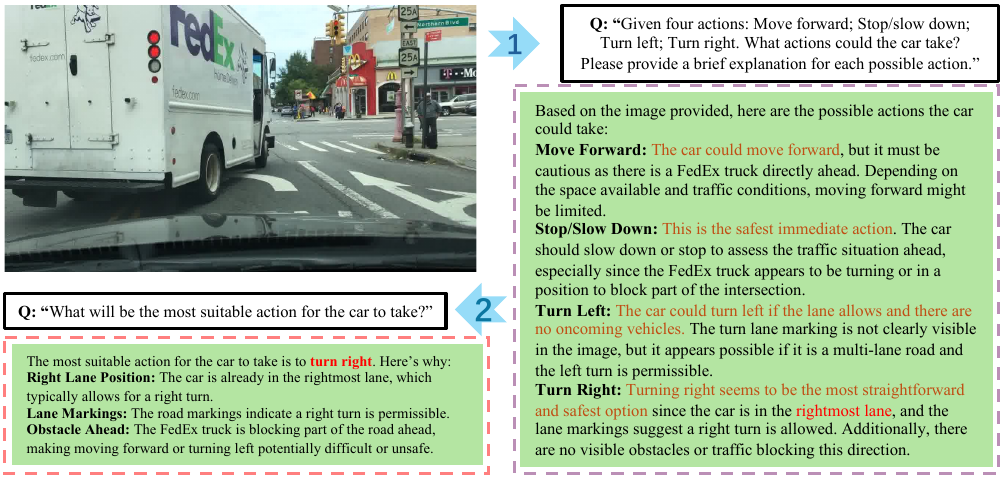}
    \vspace{-7mm}\caption{The failure case of GPT-4V for driver decision-making demonstrates that GPT-4V tends to generate vague and wrong answers for decision-making. It generates all possible action choices which will lead to system confusion. In this case, GPT-4V mistakenly recognizes the driving car in ``the rightmost lane'' which finally generates the wrong decision.}
    \label{gpt-error}\vspace{-5mm}
\end{figure*}

Recently, vision-language models (VLMs) have been increasingly explored for integration into autonomous driving systems \cite{ma2024dolphins, pan2024vlp, sima2024drivelm, tian2024drivevlm, nie2024reason2drive}. For example, VLP \cite{pan2024vlp} leverages large language models to generate text-based features aligned with visual data, achieving state-of-the-art performance in planning tasks. However, most existing approaches adopt LLaVA-style \cite{li2024llava, sima2024drivelm, tian2024drivevlm, nie2024reason2drive} methodologies, in which VLMs are prompted to produce textual outputs and natural language explanations for driving decisions, as illustrated in Figure~\ref{method-comparison}(a). We argue that such approaches overly depend on the generative capabilities of foundation models to directly make driving decisions based on handcrafted prompts, without effectively capturing fine-grained spatial relationships between local objects—an essential component of reliable and safe autonomous behavior. This concern is echoed in the comprehensive study by \cite{wen2023road}, which shows that while GPT-4V excels in general scene understanding, it frequently overlooks critical local details, resulting in substantial decision-making errors. In Figure~\ref{gpt-error}, we provide an example of GPT-4V generating vague and unreliable responses in a driving scenario.


Consequently, a natural question arises: ``\emph{How can we effectively leverage VLMs for human-like decision-making in autonomous driving?}'' While current approaches face limitations, the potential of VLMs to enhance decision-making precision remains promising. For instance, \cite{wen2023road} showcases the visual reasoning capabilities of GPT-4V through compelling qualitative examples. Moreover, our own experiments confirm that GPT-4V demonstrates strong proficiency in describing visual scenes using natural language. As illustrated in Figure~\ref{fig:chatgpt}, with carefully designed prompts, VLMs can translate visual signals into human-understandable descriptions of the surrounding environment. This capability motivates us to explore the use of VLMs for generating linguistic scene descriptions, which can provide rich semantic context and serve as a complementary modality to visual data, ultimately supporting more interpretable and accurate decision-making.

To address this question, the core motivation of our work is to leverage the strong scene reasoning capabilities of VLMs while mitigating their limitations in handling high-level reasoning. Specifically, we aim to use VLMs not as direct decision-makers, but as semantic enhancers that enrich scene understanding through natural language descriptions grounded in visual context. We employ state-of-the-art VLMs such as GPT-4V to generate high-level linguistic descriptions using instruction-based prompting for existing vision-based autonomous driving datasets. However, since VLMs often hallucinate or misrepresent the locations of local objects, we incorporate Grounding DINO \cite{liu2023grounding} to accurately detect and localize visual entities based on object-induced textual prompts. This alignment between structured visual inputs and GPT-4V-generated descriptions, illustrated in Figure~\ref{method-comparison}(b), allows us to associate semantic concepts with precise visual regions. Building on this, we design a new dual-branch decision-making framework that captures more detailed visual cues and models their intra- and inter-modal interactions, thereby supporting more accurate and interpretable driving decisions.
To sum up, our contributions are highlighted as follows:
\begin{itemize} \item First, we evaluate state-of-the-art VLMs (e.g., GPT-4V) in direct decision-making tasks and highlight their limitations. Based on this insight, we propose a new perspective that leverages VLMs to enrich vision-based autonomous driving datasets with semantically meaningful language descriptions through instruction-based prompting, grounded in local visual object relationships.
\item Second, we introduce a dual-branch multi-modal cross-attention network that effectively captures complementary information between visual and linguistic modalities. Additionally, we incorporate multi-instance learning to identify salient visual and textual concepts, enabling interpretable and context-aware decision-making.
\item Finally, we propose a post-hoc decision refinement strategy that utilizes VLMs without model retraining. Extensive quantitative and qualitative experiments on two autonomous driving benchmarks validate the effectiveness of our framework and demonstrate the advantages of our enriched multimodal data over existing methods.
\end{itemize}

\section{Related Works}
\label{sec:related}




The recent rapid development of Large Models has spurred numerous attempts to develop applications in the domain of autonomous driving \cite{atakishiyev2021explainable, keysan2023can}. Some researchers have attempted to leverage the decision-making capabilities of large models for planning in autonomous driving. MTD-GPT \cite{liu2023mtd} utilizes a sequence modeling approach and fine-tuning for decision-making problems at intersections. Similarly, DiLu \cite{wen2023dilu} employs a Large Vision Language Model agent to solve decision-making with clear prompt design, achieving comparable performance with state-of-the-art RL-based methods. \cite{cui2024receive} adopts a prompt-based approach with VLM to carry out decision-making on the highway. For data contribution, DriveLM \cite{sima2024drivelm} proposes a popular Visual Question Answer(VQA) dataset for multiple tasks in the autonomous driving area. Additionally, there are attempts with such Visual-Question-Answering tasks like DriveGPT4 \cite{xu2023drivegpt4}, VLAAD \cite{park2024vlaad}, and Dolphins \cite{ma2024dolphins}. However, the above methods heavily rely on the decision-making abilities of VLMs but ignore one significant issue-``\textit{Can Large Vision Langugae Models directly make reliable decisions like humans?}'' The results in Figure \ref{gpt-error} clearly state that GPT-4V is not reliable for decision-making in complex scenarios. Moreover, these approaches are unable to provide interpretations for the decisions made. \textbf{More related works are presented in supplementary material.}



\section{Methods}
\label{sec:method}

\subsection{Motivation}

With the rapid advancement of large Vision-Language Models (VLMs), there has been growing interest in their potential to enhance autonomous systems. However, as illustrated in Figure~\ref{gpt-error}, current VLMs \cite{sima2024drivelm, achiam2023gpt, ma2024dolphins} often struggle to consistently provide reliable decisions and grounded explanations in safety-critical domains like autonomous driving. Although they excel in global scene interpretation, their decision-making often suffers from hallucinations and a lack of attention to fine-grained, spatially localized details \cite{wen2023road}. In contrast, Figure~\ref{fig:chatgpt} demonstrates that, with carefully designed prompts, VLMs can effectively generate semantically rich descriptions of a scene, highlighting their potential as auxiliary tools in interpreting complex driving environments.

These observations motivate us to further investigate how VLMs can meaningfully contribute to driver decision-making. Rather than relying on VLMs for direct decision outputs, we propose to harness their high-level semantic understanding to enrich visual signals with complementary textual scene descriptions. This approach aims to bridge the semantic gap between raw sensory input and reasoning by introducing structured, interpretable information into the decision pipeline, thus supporting more transparent and context-aware autonomous driving systems.

\subsection{VLM Enriched BDD-OIA Benchmarks}

\begin{figure}[t]
    \centering
    \includegraphics[width=1\linewidth]{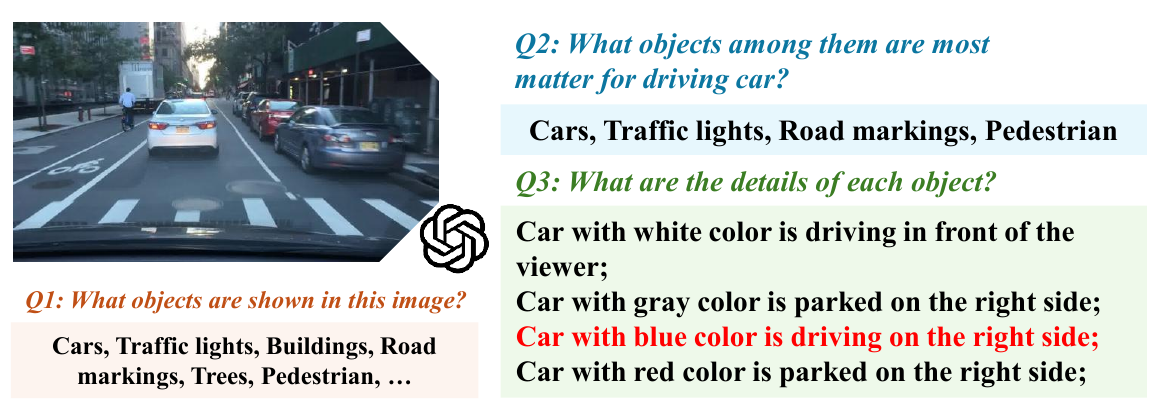}
    \vspace{-6mm}\caption{Overview of GPT-4V preprocessing with three questions in Chain-of-Thought design.}
    \label{fig:chatgpt}\vspace{-4mm}
\end{figure}

\begin{figure}[t]
    \centering
    \includegraphics[width=1\linewidth]{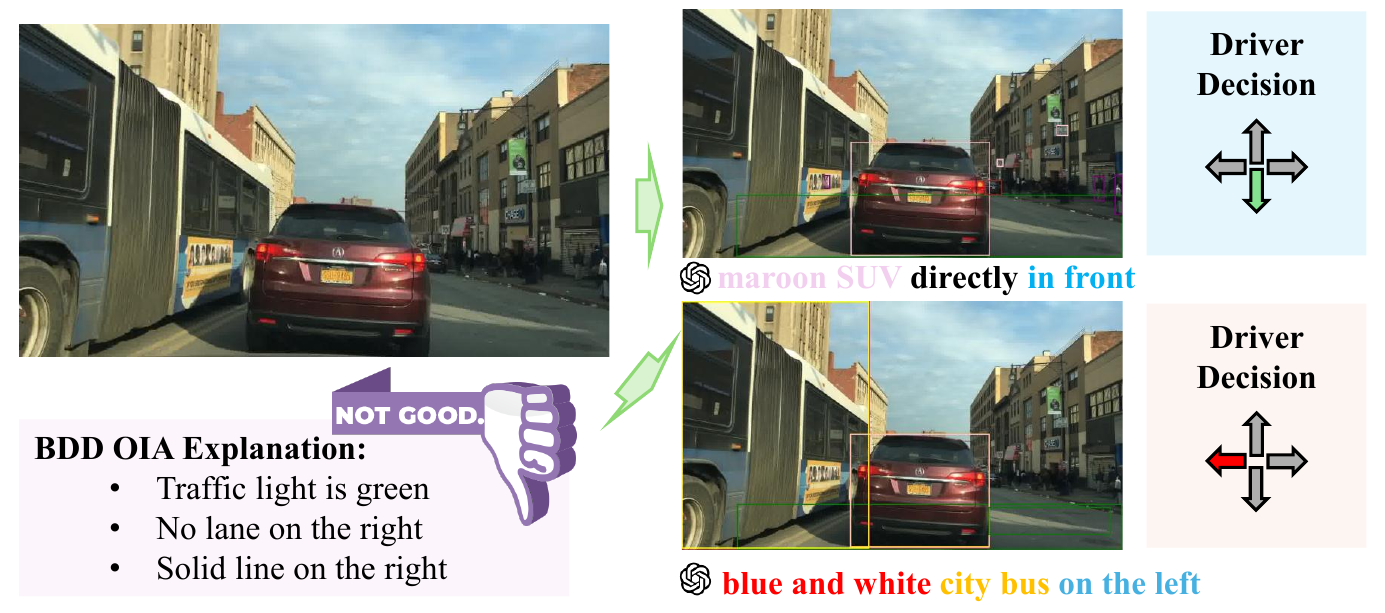}
    \vspace{-6mm}\caption{Examples of enriched descriptions from GPT-4V on BDD-OIA dataset. The first column shows the original BDD-OIA explanations have the wrong explanation(``Traffic light is green'') and missing details. The supplementary descriptions from GPT-4V in the second column demonstrate the reasons why ``stop/slow down'' and not ``left turn''.}\vspace{-4mm}
    \label{fig:oia}
\end{figure}

BDD-OIA \cite{xu2020explainable} is a widely used interpretable autonomous driving dataset, derived from BDD100K \cite{yu2020bdd100k}, which provides 21 human-labeled attribute-based explanations for driver decisions. However, these explanations are often vague or inaccurate \cite{bddoia}, leading to ambiguity and confusion in many scenarios. For instance, labels such as ``follow traffic'' on a congested street or ``traffic light'' without further clarification fail to offer meaningful guidance. As illustrated in Figure~\ref{fig:oia}, one example from BDD-OIA describes the scene with ``traffic light is green,'' even though no visible traffic light appears in the image. Moreover, it lacks justification for critical decisions like ``stop/slow down'' or avoiding a ``left turn.'' Another major limitation is the absence of visual object annotations, such as bounding boxes and semantic labels, due to the high cost of manual annotation. To address these issues, we propose using advanced vision-language models (VLMs), such as GPT-4V and Qwen2.5-VL, to automatically generate enriched and calibrated scene descriptions. As shown in the second column of Figure~\ref{fig:oia}, these model-generated descriptions offer richer context and more interpretable cues, which are crucial for understanding and justifying driver decisions.

As shown in Figure~\ref{fig:chatgpt}, we design targeted instruction prompts for GPT-4V with an emphasis on decision-critical details, such as road conditions, pedestrian behaviors, and nearby dynamic objects. Since GPT-4V processes only a single image per prompt, we follow the common practice in prior works \cite{bddoia, jing2022inaction} and select the last frame of each driving video clip—this frame is typically the most indicative for decision-making. To enrich the dataset while mitigating hallucination issues commonly observed in VLM outputs, we adopt a cascade structure inspired by Chain-of-Thought Prompting \cite{wei2022chain}. Specifically, our pipeline involves three sequential steps: 

\noindent\ding{182} We input the last frame of each video clip into GPT-4V with ``\textit{Q1: What objects are shown in this image?}''; 

\noindent\ding{183} we follow up the response to ask, ``\textit{Q2: Which objects among them are most matter for driving car?}''; 

\noindent\ding{184} we obtain the detailed descriptions from GPT-4V of each relevant object identified in the second step by following the prompt ``\textit{Q3. What are the details of each object?}''. 

\begin{figure}[t]
    \includegraphics[width=1\linewidth]{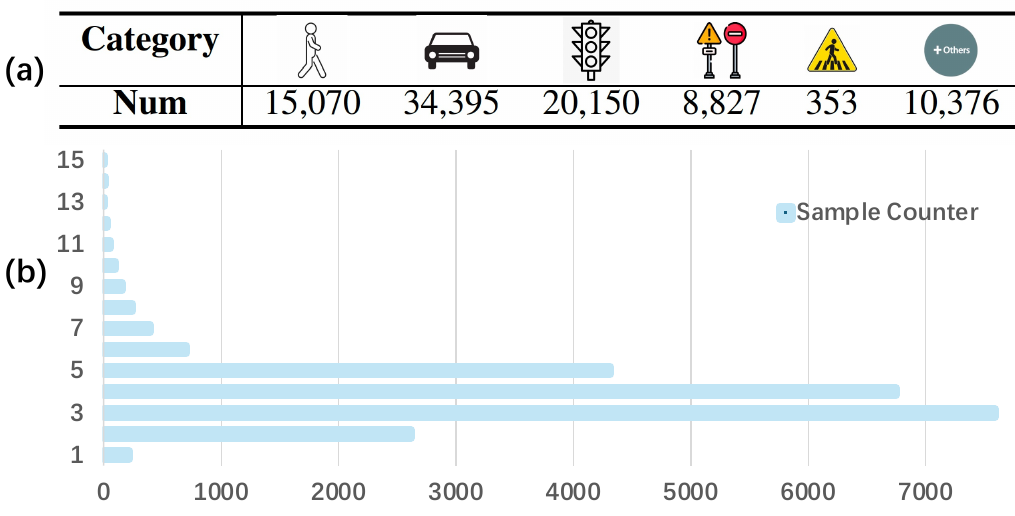}
        \vspace{-5mm}\caption{(a) Sample size per super category (b) Distribution of annotation number per sample on enriched BDD-OIA.}\vspace{-5mm}
        \label{fig:statistic}
\end{figure}

This helps filter out noise and reduces hallucinations, improving the quality of the generated data. We report a statistic summary in Figure \ref{fig:statistic}(a), where we group all fine-grained categories generated by GPT-4V into six supercategories (details in supplementary materials). We also provide the distribution of the annotation number per sample (Figure \ref{fig:statistic}(b)) with $\le$5 GPT-4V annotations for most cases.

\subsection{Dual-Branch Multi-modal Decision Model}

In this section, we propose a dual-brach multi-modal framework, which combines the explainable textual descriptions generated by VLM preprocessing with traditional visual data to enhance decision-making in autonomous driving. Given a driving scene video clips set $\{\mathcal{X}_{i}\}_{i=1}^{N}$, where $N$ is the number of video clips with $t$ frames and $i$ is the index of the sample. The decision-making label is associated as $\{\mathbf{y}_{i}\}_{i=1}^{N}$, where $\textbf{y}_{i} \in \mathbb{R}^{C}$, $C$ is the total number of decision categories, and $\mathbf{y}_{i,c}$ indicates the existence of decision $c$ in $\mathcal{X}_{i}$. Note that $\mathbf{y}_{i}$ becomes a multi-label vector when there is more than one decision in the input data. As described before, we also generate textual scene descriptions set \(\left\{ d_i^1, d_i^2, \ldots, d_i^s \right\}\) $\in T$ for $i$-th sample, in which $s$ is the number of textual descriptions for each case.

\subsubsection{Multi-modal Feature Extraction}

\begin{figure*}[t]
    \centering
    \includegraphics[width=1\linewidth]{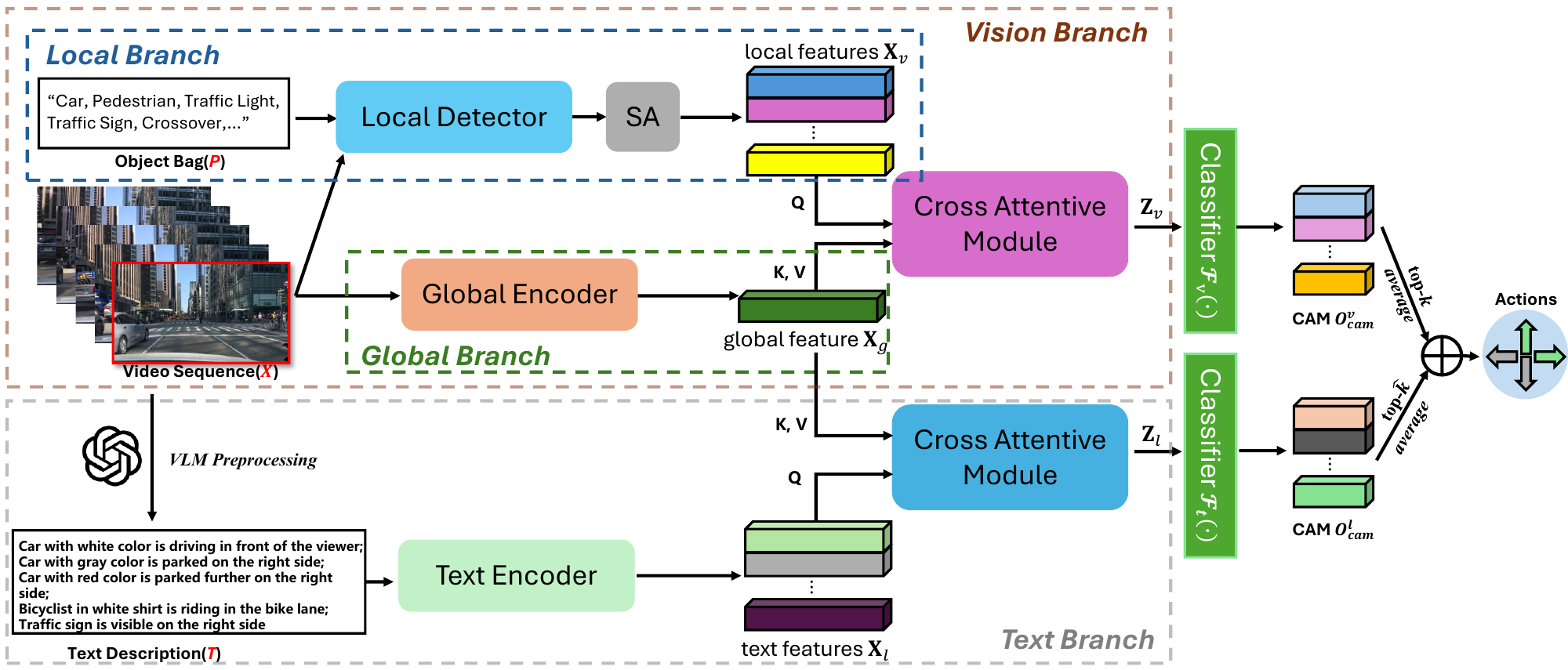}
    \vspace{-6mm}\caption{Framework of visual and textual concepts fusion for final decision-making. In the vision branch, local encoder matches object bag $P$ with images to generate local features for each category, which correlate with global feature to capture salient visual concepts. In the text branch, the descriptions from GPT-4V align with global feature for salient textual concepts. }
    \label{fig:framework}\vspace{-3mm}
\end{figure*}

As depicted in Figure \ref{fig:framework}, we propose a dual-branch multi-modal framework to explore the complementary information between the visual and the language modalities. Firstly, we introduce one global feature encoder to extract the global scene features $\mathbf{X}_{g} \in \mathbb{R}^{t \times  D}$ for the whole video clip, where $t$ is the length of each video clip and $D$ is the channel dimension. This provides a comprehensive understanding of the overall scene, capturing context such as road conditions and object motions.

However, relying solely on global features can miss critical details and social relations about specific objects (like most VLMs), which are vital for driving decisions. To capture local objects, existing methods \cite{jing2022inaction, xu2020explainable} utilize ROI (Region of Interest) features from object detection or localization networks such as FastRCNN \cite{girshick2015fast} or YOLO \cite{redmon2016you}. Although effective, these methods often include background noise, and are difficult to track the most relevant objects.




To address this issue, we propose to focus on local objects that are most mentioned in the textual descriptions generated by GPT-4V. We count the object category frequency in the text descriptions with automatic script before training and collect the top 10 frequent class as ``object bag'' \(\left\{ \text{obj}_1, \text{obj}_2, \ldots \right\}\) $\in P$. Note that different datasets have different $P$. As shown in Figure \ref{fig:framework}, we use $P$ as text prompts and input it into the local detector such as grounding DINO along with the last frame of video clips (similar to text generation) to extract local visual features $\mathbf{X}_{v} \in \mathbb{R}^{n \times D}$ , where $n$ is the total number of identified objects. Note that the reason why we use $P$ instead of $T$ as text prompts is that we want to detect all relevant visual objects while the textual descriptions from GPT-4V would miss some objects. This method effectively filters out background noise and makes it easier to search and track the most salient objects, thereby improving the precision and reliability of the system.



While the local object detector effectively tracks relevant objects based on simple text categories, it cannot match the specific object from the detailed semantic information provided by GPT-4V. To address this, we introduce a text branch to encode textual features $\mathbf{X}_{l} \in \mathbb{R}^{s \times D}$ from scene descriptions $T$ generated by GPT-4V. The text branch complements the vision branch by capturing nuanced details for the local objects such as ``car with red brake light''. However, due to the limitations of GPT-4V, the language information may not cover all local objects identified in the vision branch, leading to incomplete semantic information for some objects, as mentioned before. Despite this limitation, the enriched textual descriptions significantly enhance the overall scene understanding by providing additional semantic information.

\subsubsection{Dual-branch Local-Global Interactive Fusion}

To efficiently explore the complementary information between vision and language modality, we develop a dual branch multi-modal architecture with a cross-attention Transformer to explore the vision-vision and vision-language relationships. In this manner, the most salient parts are addressed and the redundant noise will be filtered out by such a mechanism. 

Generally, both the local vision features $\mathbf{X}_{v}$ and textual features $\mathbf{X}_{l}$ provide detailed information for specific objects while global feature $\mathbf{X}_{g}$ represents the whole video. It would be better to use $\mathbf{X}_{v}$ and $\mathbf{X}_{l}$ as queries to trigger the salient area in the global feature. In addition, it is critical to build the relationship between local vision features. Considering this, we propose to utilize the cross-attentive module to serve as a bridge for the feature fusion:
\begin{equation}
\label{transformer equation}
\left\{
\begin{aligned}
& \mathbf{Z}_{v} = \mathcal{F}_\mathrm{vision}(\mathbb{SA}(\mathbf{X}_{v}), \mathbf{X}_{g}), \\
& \mathbf{Z}_{l} = \mathcal{F}_\mathrm{text}(\mathbf{X}_{l}, \mathbf{X}_{g}),
\end{aligned}
\right.
\end{equation}
where $\mathcal{F}_\mathrm{vision}(\cdot)$ and $\mathcal{F}_\mathrm{text}(\cdot)$ are cross-attentive modules in the vision and text branch, respectively. $\mathbb{SA}(\cdot)$ denotes self-attention module. $\mathbf{Z}_{v} \in \mathbb{R}^{n \times D}$ and $\mathbf{Z}_{l} \in \mathbb{R}^{s \times D}$ are refined output features for vision and text branch. More details are shown in supplementary materials.

\subsubsection{Decision via Multi-Instance Learning}
Here, we have obtained the refined features $\mathbf{Z}_{v}$ and $\mathbf{Z}_{l}$ for two branches. It is significant for safety concerns to capture the salient objects and textual descriptions in the input data as explanations of decision-making. However, most datasets do not have such labels and it is time-consuming and expensive for human annotations. Inspired by weakly supervised setting \cite{hu2025enhancing, hu2024weakly}, we propose to introduce a multi-instance learning loss for model training instead of using the conventional multi-label loss in BDD-OIA \cite{xu2020explainable}. As shown in Figure \ref{fig:framework}, we develop separate classifiers $\mathcal{F}_\mathrm{v}(\cdot)$ and $\mathcal{F}_\mathrm{t}(\cdot)$ for each branch and use the Class Activation Matrix (CAM) for the final prediction, indicated as $\mathbf{O}_\mathrm{cam}^{v} \in \mathbb{R}^{n \times C}$ and $\mathbf{O}_\mathrm{cam}^{l} \in \mathbb{R}^{s \times C}$ respectively. The prediction of the final decision $\mathbf{O}_\mathrm{pred} \in \mathbb{R}^{C}$ would be the weighted average sum between $\mathbf{O}_\mathrm{cam}^{v}$ and $\mathbf{O}_\mathrm{cam}^{l}$ as:
\begin{equation}
    \mathbf{O}_\mathrm{pred} = \lambda\mathcal{F}_\mathrm{AVG}(\mathcal{F}_\mathrm{k}(\mathbf{O}_\mathrm{cam}^{v})) + (1-\lambda)\mathcal{F}_\mathrm{AVG}(\mathcal{F}_\mathrm{\hat{k}}(\mathbf{O}_\mathrm{cam}^{l})),
\end{equation}
where $\lambda$ is the hyperparameter, $\mathcal{F}_\mathrm{AVG}(\cdot)$ is the average pooling function along the sequence dimension, $\mathcal{F}_\mathrm{k}(\cdot)$ and $\mathcal{F}_\mathrm{\hat{k}}(\cdot)$ are different top-$K$ functions for vision and text branch. The final loss will be set as:
\begin{equation}
    \mathcal{L} = \mathcal{L}_\mathrm{MIL}(\mathbf{O}_\mathrm{pred}, \mathbf{y}_{i}).
\end{equation}

\begin{figure}[t]
    \centering
    \includegraphics[width=1\linewidth]{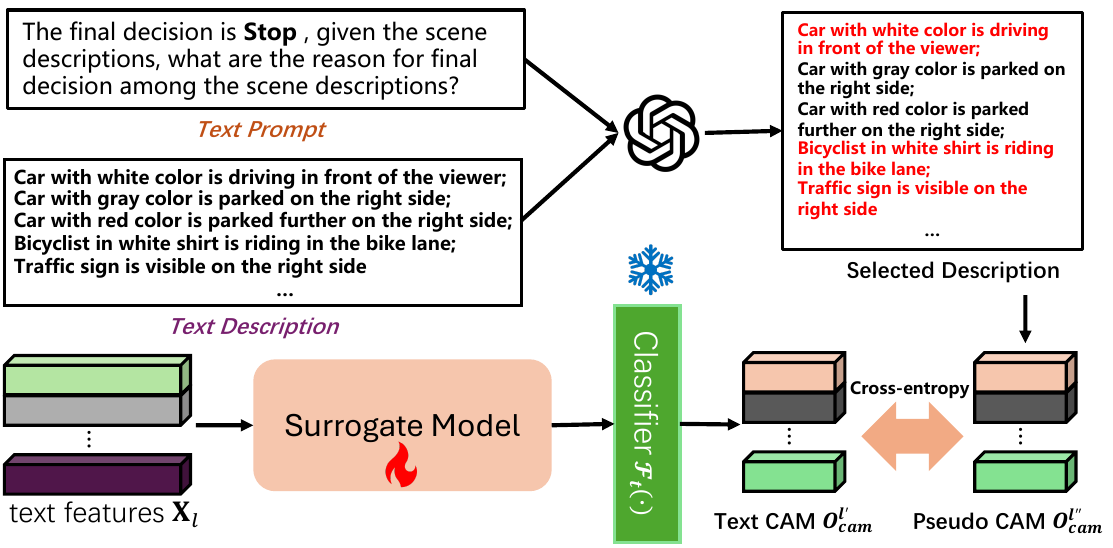}
     \vspace{-5mm}\caption{Mechanism of refinement with VLM. The text descriptions($T$) and decision-guided text prompts are input into GPT-4V to match decision label with related text descriptions. The selected descriptions will set as pseudo labels for training surrogate model to refine text CAM.} \vspace{-4mm}
    \label{fig:refinement}
\end{figure}

\subsubsection{Further Refinement with VLMs}
The fusion of vision and language information achieves better performance than only a single modality, as shown in Tables 1 and 2 in supplementary material. However, we notice that the text branch might not fully utilize the language information, since $\mathbf{Z}_{l}$ is mainly from the global feature $\mathbf{X}_{g}$. This motivates us to further optimize the text branch with $\mathbf{X}_{l}$. Although VLMs have limited decision-making ability with only visual input, it is still desirable to do correct reasoning with detailed prior knowledge \cite{wen2023road}. As illustrated in Figure \ref{fig:refinement}, we input the enriched text descriptions and the final decision label into VLM to conduct reasoning for the decision label with related text descriptions on the training data. After getting related descriptions for each decision, we transfer such knowledge to pseudo labels as $\mathbf{O}_\mathrm{cam}^{l''}$. For the refinement module, we introduce a simple surrogate model to build the bridge between $\mathbf{X}_{l}$ and trained classifier $\mathcal{F}_\mathrm{t}(\cdot)$. The generated CAM $\mathbf{O}_\mathrm{cam}^{l'}$ is supervised by pseudo CAM $\mathbf{O}_\mathrm{cam}^{l''}$ with cross-entropy loss. Note that we only train the surrogate model and other modules which are trained before will be frozen. The refined output will be $\mathbf{O}_\mathrm{cam}^{\hat{l}} = \frac{1}{2}(\mathbf{O}_\mathrm{cam}^{l} + \mathbf{O}_\mathrm{cam}^{l'})$ and $\mathbf{O}_\mathrm{pred} = \lambda\mathcal{F}_\mathrm{AVG}(\mathcal{F}_\mathrm{k}(\mathbf{O}_\mathrm{cam}^{v})) + (1-\lambda)\mathcal{F}_\mathrm{AVG}(\mathcal{F}_\mathrm{\hat{k}}(\mathbf{O}_\mathrm{cam}^{\hat{l}}))$.
\section{Experiments}
\label{sec:experiment}

\subsection{Experimental Settings}

\noindent\textbf{BDD-OIA dataset} \cite{xu2020explainable} is a subset of BDD100K \cite{yu2020bdd100k} consisting of 22,924 5-second video clips, which are annotated with 4 action decisions. We simplify the decision label ``move forward'' as \textbf{F}, ``stop/slow down'' as \textbf{S}, ``left turn'' as \textbf{L}, ``right turn'' as \textbf{R} in Table \ref{bdd-oia}. The original protocol of BDD-OIA adopts the last frame as visual input. We extend the last frame into 15-frames video clips captured from BDD100k to be consistent with current decision-making methods. The resolution of each frame is 1280 $\times$ 720. As there are multiple possible action choices for each sample, we evaluate the performance by F1 score for each specific action, overall F1 score, and the class-wise average F1 score for action decision-making.

\begin{table*}[t]
\centering
\caption{Performance comparison of different methods on BDD-OIA dataset. Note that all VLMs are tested without finetuning on BDD-OIA, and others are finetuned with BDD-OIA.}\vspace{-2mm}
\begin{tabular}{c|cccccc}
\Xhline{1pt}
 \textbf{Method}             & \textbf{F}     & \textbf{S}     & \textbf{L}     & \textbf{R}     & \textbf{F1$_{all}$} & \textbf{mF1} \\ \hline
 Dolphins-7B \cite{ma2024dolphins}       & 0.713          & 0.244          & 0.09          & 0.12          & 0.447                    & 0.241             \\
                                 LLAVA-NEXT-8B \cite{li2024llava}              & 0.700          & 0.522          & 0.232          & 0.237          & 0.502                    & 0.423             \\
                                CogVLM2-19B \cite{wang2023cogvlm}                & 0.700          & 0.621          & 0.384          & 0.402          & 0.538                    & 0.527             \\
                                InternVL2-26B \cite{chen2023internvl}       & 0.704          & 0.591          & 0.427          & 0.455          & 0.544                    & 0.526             \\
                                InternVL2-40B \cite{chen2023internvl}       & 0.705          & 0.625          & 0.438          & 0.468          & 0.563                    & 0.558             \\ \hline
 ResNet \cite{xu2020explainable}                     & 0.755          & 0.607          & 0.098          & 0.108          & 0.601                    & 0.392             \\
                                OIA \cite{xu2020explainable}                         & 0.829          & 0.781          & 0.630          & 0.634          & 0.734                    & 0.718             \\
                                InAction (proposals-based) \cite{jing2022inaction}        & 0.795          & 0.743          & 0.597          & 0.613          & 0.706                    & 0.687             \\
                                InAction (global) \cite{jing2022inaction}           & 0.800          & 0.747          & 0.612          & 0.619          & 0.714                    & 0.694             \\ \cline{1-7} 
                                Ours (last frame) & 0.838 & 0.803 & 0.655 & 0.659          & 0.750           & 0.738    \\ 
                                Ours (video) & \textbf{0.846} & \textbf{0.812} & \textbf{0.663} & \textbf{0.671}          & \textbf{0.762}           & \textbf{0.748}    \\\Xhline{1pt}
\end{tabular}\vspace{-3mm}
\label{bdd-oia}
\end{table*}

\noindent\textbf{PSI dataset} \cite{chen2021psi} consists of videos with complex environments. Compared with the old PSI dataset in \cite{jing2022inaction}, the new PSI dataset includes more videos and annotations. It has 22,579 2-second video clips, which are annotated with 3 single-label decisions (``increase speed'', ``decrease speed'', and ``maintain speed''). We adopt ``+'', ``-'', ``$\circ$'' to represent ``increase speed'', ``decrease speed'', and ``maintain speed'' in Table \ref{psi}. PSI dataset already has human-annotated language descriptions for each video clip, which is used as the enriched description. We take the same metrics as BDD-OIA to evaluate the model performance on PSI.



\noindent\textbf{Implementation Details}: We follow existing works \cite{jing2022inaction, xu2020explainable} and utilize the FastRCNN \cite{girshick2015fast} pretrained on BDD-100k to extract the global feature of each sample. The feature output by FastRCNN is a sequence of feature maps whose size is 15 $\times$ 1024 $\times$ 45 $\times$ 80. We apply two downsample 3D convolution layers followed by average pooling to transfer the global feature to a sequence of 256-dimensional vectors. For local features, we exploit the potential of the grounding DINO \cite{liu2023grounding} with three downsample convolution layers to generate local tokens for the input text prompt. Each object token is a 256-dimensional vector. We utilize CLIP \cite{radford2021learning} text encoder to extract the text feature, which is a 1024-dimensional vector, and introduce two convolution layers to downsample the size to 256. To make the input consistent, we fix the number of $n$ as 80 and $s$ as 20 for the BDD-OIA dataset, and for the PSI dataset, we set the number of $n$ as 40 and $s$ as 24. For the cross-attentive module, we adopt a multi-head transformer with the number of heads as 8. Every transformer has 1 attention layer with a dimension of 256. The classification module contains 3 convolution layers, between which two dropout layers with a rate of 0.7 are added to regularize intermediate features. The surrogate model includes three convolution layers with ReLU activation. For training, we first train the whole framework in Figure \ref{fig:framework} and then train the refinement module with a frozen classifier. For inference, the model outputs the final decision with the refinement module.

For hyperparameters, we train the model with the learning rate of 1e-4 by using Adam optimizer \cite{kingma2014adam}, and we set the batch size at 128 and the training epochs at 100 for all parts. We choose $k$ = 16 and $\hat{k}$ = 1 for the vision and text branch respectively in BDD-OIA dataset. For PSI dataset, we select $k$ = 8 and $\hat{k}$ = 1 separately. Additionally, we set $\lambda$ as 0.8 for both datasets. All experiments are conducted on a single RTX 6000Ada GPU.

\subsection{Comparison Results}


In Table \ref{bdd-oia}, we first evaluate zero-shot decision making of several VLMs on BDD-OIA. Among them, LLAVA-NEXT-8B \cite{li2024llava}, CogVLM2-19B \cite{wang2023cogvlm} and Dolphins \cite{ma2024dolphins} adopt FP16 for inference, while InternVL2-26B and InternVL2-40B \cite{chen2023internvl} utilize INT8 due to hardware limitations. For all tested VLMs, we utilize video clips as input data. Despite this, these VLMs still couldn't beat finetuned ResNet on BDD-OIA even though they are pretrained by large scale datasets. Additionally, it is worth noting that Dolphins \cite{ma2024dolphins}, a model pretrained for multiple driving tasks on BDD-X \cite{kim2018textual} (a superset of BDD-OIA, including BDD-OIA but much larger), performs worse than other VLMs in Table \ref{bdd-oia}—let alone when compared with our proposed method. This highlights a critical issue: \emph{even VLMs pretrained on BDD-OIA like data struggle to achieve competitive performance, underscoring the effectiveness and robustness of our approach.} It also supports the conclusion in \cite{wen2023road} that VLMs exhibit a limited capacity of decision making.



For the non-VLM methods in Table \ref{bdd-oia}, we observe that our method outperforms all results in all metrics and establishes a new state-of-the-art performance on BDD-OIA with mean F1 score of 74.8\% and general F1 score of 76.2\%. Compared to the original BDD-OIA \cite{xu2020explainable} method, our method achieves average improvements of 4\% on all four actions. For the model which provides interpretability (InAction \cite{jing2022inaction}, etc.), our model also outperforms with average 7.8\% on four actions. In addition, our network achieves 6.7\% and 7.8\% improvements to InAction on average F1 score and the overall F1 score, respectively. Since the original BDD-OIA protocol uses the last frame as the visual input, we also evaluate our method under this setting. However, we observe consistent improvements across all metrics when using full video input compared to a single frame, further demonstrating the importance and effectiveness of temporal information in driving decision-making.

\begin{table}[t]
\centering
\caption{Performance comparison on PSI dataset}\vspace{-3mm}
\scalebox{0.92}{
\begin{tabular}{c|p{5mm}p{5mm}p{5mm}cc}
\Xhline{1pt}
\textbf{Method}                                        & \textbf{+} & \textbf{$\circ$} & \textbf{-} & \textbf{F1$_{all}$} & \textbf{mF1} \\ \hline
OIA (global) \cite{xu2020explainable} & 0.348              & 0.411                   & 0.133              & 0.357                    & 0.297             \\
OIA \cite{xu2020explainable}          & 0.459              & 0.534                   & 0.17               & 0.416                    & 0.387             \\ \hline
Ours (last frame)                                     & 0.491              & 0.608                   & 0.385              & 0.554                    & 0.495             \\
Ours (video)                            & \textbf{0.501}     & \textbf{0.627}                   & \textbf{0.396}     & \textbf{0.571}           & \textbf{0.508}    \\ \Xhline{1pt}
\end{tabular}
}\vspace{-4mm}
\label{psi}

\end{table}

\begin{table}[t]
\centering
\caption{Ablation study of different modules on BDD-OIA dataset}\vspace{-3mm}
\scalebox{0.9}{\begin{tabular}{p{1mm}p{1mm}p{1mm}|p{5mm}p{5mm}p{5mm}ccc}
\Xhline{1pt}
\multicolumn{3}{c|}{\textbf{Modules}}                 & \multirow{2}{*}{\textbf{F}} & \multirow{2}{*}{\textbf{S}} & \multirow{2}{*}{\textbf{L}} & \multirow{2}{*}{\textbf{R}} & \multirow{2}{*}{\textbf{F1$_{all}$}} & \multirow{2}{*}{\textbf{mF1}} \\ \cline{1-3}
\textbf{V} & \textbf{T} & \textbf{R} &                             &                             &                             &                             &                                           &                                    \\ \hline
                &               &                     & 0.813                       & 0.739                       & 0.595                       & 0.607                       & 0.705                                     & 0.689                              \\
       $\checkmark$         &               &                     & 0.826                       & 0.78                        & 0.641                       & 0.659                       & 0.738                                     & 0.726                              \\
            &     $\checkmark$          &                     & 0.811                       & 0.771                       & 0.626                       & 0.637                       & 0.722                                     & 0.711                              \\
         $\checkmark$       &      $\checkmark$         &                     & 0.830                       & 0.791                       & 0.645                       & 0.660                       & 0.743                                     & 0.732                              \\
          $\checkmark$      &      $\checkmark$         &        $\checkmark$             &  \textbf{0.846}              & \textbf{0.812}              & \textbf{0.663}              & \textbf{0.671}              & \textbf{0.762}                            & \textbf{0.748}                     \\ \Xhline{1pt}
\end{tabular}}\vspace{-3mm}
\label{ablation-bdd}
\end{table}

\begin{figure*}[t]
    \centering
    \includegraphics[width=1\linewidth]{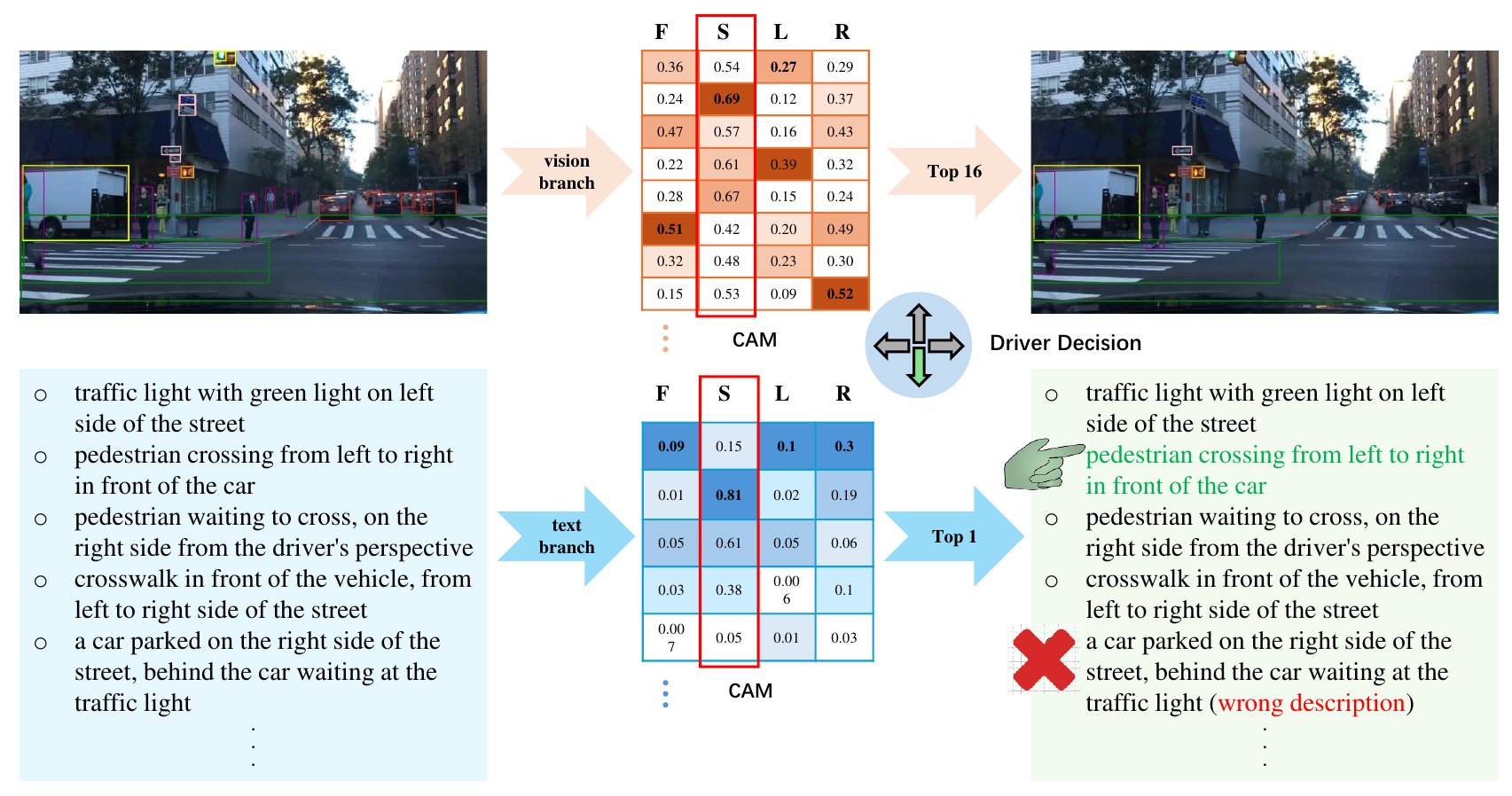}
    \vspace{-7mm}\caption{Visual and language explanations of the last frame on BDD-OIA dataset, where the first column shows all local objects identified with object bag $P$ and text descriptions $T$ from GPT-4V, the second column represents the intermediate CAM results in different branches, the last column reports our salient visual concepts, text concepts, and the driver decision. }
    \label{fig:oia_rs}\vspace{-4mm}
\end{figure*}

Table \ref{psi} presents the results on the PSI dataset, where we retrain BDD-OIA \cite{xu2020explainable} under the new setting. Due to the imbalanced distribution in PSI—particularly the significantly lower number of samples in the ``Decrease Speed'' category—OIA models exhibit noticeably worse performance in this class. Interestingly, our method achieves superior performance on the ``Decrease Speed'' decision, with an average accuracy of 39.6\%, showcasing its robustness in imbalanced scenarios. Similar to our evaluation of the BDD-OIA dataset, we also assess performance using only the last frame as input. The results again confirm that incorporating temporal information leads to more accurate prediction.

\subsection{Ablation Study}

Table \ref{ablation-bdd} presents the ablation study of different modules on the BDD-OIA dataset. We use \textbf{V} to denote the vision branch, \textbf{T} for the text branch, and \textbf{R} for the refinement module. The first row in Table \ref{ablation-bdd} represents the baseline model, which only employs a global branch for decision prediction. The results show that relying solely on global features fails to capture the fine-grained details necessary for accurate decision-making, highlighting the importance of incorporating local features. While the vision branch (\textbf{V}) outperforms the text branch (\textbf{T}) individually—suggesting that GPT-4V-generated descriptions still contain some noise—combining the two modalities leads to a significant performance gain, demonstrating their complementary nature. Furthermore, the addition of the refinement module (\textbf{R}) results in further improvements, confirming its effectiveness in enhancing prediction accuracy.

Table \ref{ablation-psi} is the ablation study on the PSI dataset. Similar to Table \ref{ablation-bdd}, \textbf{V} represents the vision branch, \textbf{T} is for the text branch and \textbf{R} means refinement module. As for the effectiveness of the proposed modules, it seems that the performance of the text branch is slightly better than that in BDD-OIA. We analyze that the reason is that PSI provides reliable human-labeled descriptions, which reduce the noise compared with the text generated by GPT-4V. Additionally, our proposed dual branch still exhibits the best performance on the PSI dataset. Furthermore, the refinement module also shows the effectiveness on the PSI dataset.

\subsection{Explanation Results}
\begin{table}[t]
\centering
\caption{Ablation study of different modules on PSI dataset}\vspace{-3mm}
\scalebox{0.9}{\begin{tabular}{ccc|ccccc}
\Xhline{1pt}
\multicolumn{3}{c|}{\textbf{Modules}}                 & \multirow{2}{*}{\textbf{+}} & \multirow{2}{*}{\textbf{$\circ$}} & \multirow{2}{*}{\textbf{-}} & \multirow{2}{*}{\textbf{F1$_{all}$}} & \multirow{2}{*}{\textbf{mF1}} \\ \cline{1-3}
\textbf{V} & \textbf{T} & \textbf{R} &                             &                             &                             &                                           &                                    \\ \hline
                &               &                     & 0.403                       & 0.439                       & 0.22                        & 0.404                                     & 0.355                              \\
      $\checkmark$          &               &                     & 0.437                       & \textbf{0.63}               & 0.211                       & 0.505                                     & 0.424                              \\
                &       $\checkmark$        &                     & 0.446                       & 0.541                       & 0.353                       & 0.484                                     & 0.447                              \\
         $\checkmark$       &      $\checkmark$         &                     & 0.476                       & 0.585                       & 0.374                       & 0.531                                     & 0.478                              \\
         $\checkmark$       &     $\checkmark$          &         $\checkmark$            & \textbf{0.501}              & 0.627                       & \textbf{0.396}              & \textbf{0.571}                            & \textbf{0.508}                     \\ \Xhline{1pt}
\end{tabular}}
\label{ablation-psi}
\vspace{-6mm}
\end{table}
We aim to identify the most representative visual and textual concepts of the last frame to gain a deep understanding of what should be recognized as the main factor in related action decisions. As described in the method part, we utilize the CAM to select top-contributed local objects and text descriptions for different action decisions. The detailed results are shown in Figure \ref{fig:oia_rs}, in which the first column demonstrates all local objects detected and paired text descriptions from GPT-4V, and the second column shows intermediate CAM results of vision and text branch. Since we set $k$=16 and $\hat{k}$=1 for BDD-OIA dataset, we only show the top-16 local objects and the top-1 text description for explanations. The third column represents the action decision label as well as our explainable visual and text concepts. It is clearly observed that our framework accurately captures the representative visual object (``the pedestrian crossing the street'') and text reasoning (``pedestrian crossing from left to right in front of the car'') for the decided action (``stop/slow down''). Even for the wrong description (``a car parked on the right side of the street, behind the car waiting at the traffic light''), our model has the capability to filter it out.

\section{Conclusion}
\label{sec:conclusion}

In this work, we propose an efficient pipeline that leverages Vision-Language Models (VLMs) to enrich autonomous driving datasets with detailed semantic scene descriptions. Our dual-branch multi-modal framework models the interaction between visual and linguistic modalities, while a multi-instance learning strategy identifies salient concepts to provide interpretable decisions. A training-efficient refinement mechanism further improves decision accuracy without additional model training. Despite its effectiveness, our pipeline has limitations: VLMs may overlook certain objects, resulting in incomplete textual descriptions. Addressing this will be the focus of future work.
\clearpage
\setcounter{page}{9}
{
    \small
    \bibliographystyle{ieeenat_fullname}
    \bibliography{main}
}

\end{document}